\documentclass[11pt]{article}

\usepackage[frozencache,cachedir=.]{minted}
\usepackage{amsmath}
\usepackage{amsfonts}
\usepackage{amsthm}
\usepackage{setspace}
\usepackage{mathtools}
\usepackage{subcaption}
\usepackage{suffix}
\RequirePackage{color}
\usepackage{adjustbox}
\usepackage{hyperref}
\usepackage{authblk}
\include{math-commands}
\author[1]{Brandon T. Willard}
\affil[1]{Normal Computing}
\author[2]{Rémi Louf}
\affil[2]{Normal Computing}
\definecolor{bg}{rgb}{0.95,0.95,0.95}
\usepackage[authoryear]{natbib}
\usepackage{cleveref}
\allowdisplaybreaks
\setkeys{Gin}{keepaspectratio}
\graphicspath{{}{../figures/}{./figures/}{./}}
\setminted{breaklines=true, breakanywhere=true, breakautoindent=true}
\usepackage{adjustbox}

\newlength{\maxtabfigwidth}
\newlength{\maxtabfigheight}
\usepackage{mleftright}
\usepackage{tcolorbox}
\tcbuselibrary{minted, listings, breakable, skins}
\usepackage{algorithm}
\usepackage{algpseudocode}
\date{2023-07-14}
\title{Efficient Guided Generation for Large Language Models}
\hypersetup{
 pdfauthor={Brandon T. Willard, Rémi Louf},
 pdftitle={Efficient Guided Generation for Large Language Models},
 pdfkeywords={},
 pdfsubject={},
 pdfcreator={Emacs 30.0.50 (Org mode 9.6.7)}, 
 pdflang={English}}
\begin{document}

\newtcblisting[auto counter,number within=section]{oxtcblisting}[1]{%
	frame hidden,
	listing only,
	listing engine=minted,
	minted options={numbersep=2mm},
	breakable,
	enhanced,
	title after break={\raggedleft\lstlistingname\ \thetcbcounter~ -- continued},
	listing remove caption=false,
	arc=0pt,
	outer arc=0pt,
	boxrule=0pt,
	coltitle=black,
	colbacktitle=white,
	center title,
	#1
}
\maketitle

\begin{abstract}
In this article we show how the problem of neural text generation can be
constructively reformulated in terms of transitions between the states of a
finite-state machine. This framework leads to an efficient approach to guiding
text generation with regular expressions and context-free grammars by allowing
the construction of an index over a language model's vocabulary. The approach is
model agnostic, allows one to enforce domain-specific knowledge and constraints,
and enables the construction of reliable interfaces by guaranteeing the
structure of the generated text. It adds little overhead to the token
sequence generation process and significantly outperforms existing solutions.
An implementation is provided in the open source Python library Outlines
\citep{LoufoutlinesGenerativeModel}.
\end{abstract}

\section{Introduction}
\label{sec:intro}
We are concerned with the problem of generating sequences of tokens from a large
language model (LLM) \citep{VaswaniAttentionallyou2017,RadfordLanguagemodelsare2019} that conform to regular expressions or
context-free grammars (CFGs).  This kind of guided LLM generation is used to
make LLM model output usable under rigid formatting requirements that are either
hard or costly to capture through fine-tuning alone
\citep{Beurer-KellnerPromptingprogrammingquery2023,ScholakPICARDParsingincrementally2021,PoesiaSynchromeshReliablecode2022,RabinovichAbstractsyntaxnetworks2017,WengControllableNeuralText2021,DongCODEPGrammaticalSeq2Seq2023,PoesiaSynchromeshReliablecode2022a,GengFlexibleGrammarBasedConstrained2023,WangGrammarPromptingDomainSpecific2023}.
Such features have recently been generalized in prompting libraries and
interfaces \citep{Microsoftguidance2023,Beurer-KellnerPromptingprogrammingquery2023,RickardparserLLM2023,Rickardr2d4rellmExact2023}, but their applicability can be limited by
their scaling costs.

Most implementations of guided generation bias the score values used to
determine the probabilities of the tokens in an LLM's vocabulary.  A common and
sufficient approach involves repeated evaluations over the entire vocabulary in
order to determine which tokens are valid--according to the constraints and
previously sampled tokens--and setting the probabilities of invalid tokens to
zero.  This approach entails a fixed \(\mathcal{O}(N)\) cost for each
token generated, where \(N\) is the size of the LLM's vocabulary.

We propose an approach that uses the finite state machine (FSM) formulation of
regular expressions to both arbitrarily start and stop guided generation and
allow the construction of an index with which the set of non-zero-probability
tokens can be obtained efficiently at each step.  The result is an algorithm
that costs \(\mathcal{O}(1)\) on average.

For the regular expression case, our approach shares the most similarity with
\citet{KuchnikValidatinglargelanguage2023}, which uses a transducer
formulation to obtain FSMs defined over a language model's vocabulary, and these
FSMs contain much of the same information and scaling benefits as the indices
described here.  Our approach does not require the complete transducer
abstraction and can be used to more easily extend existing, efficient regular
expression libraries without modifying the underlying automatons and their
implementations.

More importantly, our indexing approach can also be extended to CFGs and
\(\operatorname{LALR}(1)\) parsers to allow for efficient guided generation
according to popular data formats and programming languages (e.g. JSON, Python,
SQL, etc.).  The transition to parsing is made by way of augmentations to
traditional \(\operatorname{LALR}(1)\) parser components and operations, making
it--again--an approach that can be used to extend existing parser implementations.

\section{LLM Sampling and Guided Generation}
\label{sec:org4b0d5f6}

Let \(S_t = \left(s_1 \dots s_t\right)\) represent a sequence of \(t\) tokens
with \(s_t \in \mathcal{V}\), \(\mathcal{V}\) a vocabulary, and
\(\abs{\mathcal{V}} = N\).  The vocabularies, \(\mathcal{V}\), are composed of
strings from a fixed alphabet \citep{SennrichNeuralmachinetranslation2015} and
\(N\) is often on the order of \(10^4\) or larger.

We define the next token \(s_{t+1}\) as the following random variable:
\begin{align*}
  \boldsymbol{\alpha} &= \operatorname{LLM}(S_t, \boldsymbol{\theta})\\
  s_{t+1} &\sim \operatorname{Categorical}({\boldsymbol{\alpha}})
\end{align*}

where \(\boldsymbol{\theta}\) is the set of trained parameters and
\(\boldsymbol{\alpha} \in \mathbb{R}^N\).  In the context of this paper the
function \(\operatorname{LLM}\) refers to a deep neural network trained on
next-token-completion tasks, but the method extends more generally to any
function that takes token sequences and returns a probability distribution for
the next token.

\subsection{Sampling sequences}
\label{sec:orgf466c7c}

Let \(\mathcal{F} \subset \mathcal{P}\left(\mathcal{V}\right)\), where
\(\mathcal{P}\) is the powerset operator, be subsets of multi-token strings that
end with a special token \(\texttt{EOS} \in \mathcal{V}\). The text generation
task is to draw samples from \(\mathcal{F}\).

Several procedures have been considered to generate elements of \(\mathcal{F}\).
Greedy decoding consists in generating tokens recursively, choosing the token
with highest probability at each step. Beam search also generates tokens
recursively, using a heuristic to find the mode of the distribution. More
recently, SMC sampling has also been used to generate sequences
\citep{LewSequentialMonteCarlo2023}.

\begin{algorithm}
  \caption{Basic LLM token sampling}
  \label{alg:llm-sequence-sampling}
  \begin{algorithmic}[1]
    \Function{sample\_tokens}{$L$}
      \State $\boldsymbol{s} \gets () $
      \For{$i \gets 1, L$}
        \State $\boldsymbol{\alpha} \gets $
          \Call{$\operatorname{LM}$}{$\boldsymbol{s}$, $\boldsymbol{\theta}$}
        \State Sample $s \sim \operatorname{Categorical}({\boldsymbol{\alpha}})$
        \If{$s = \texttt{EOS}$}
          \State \textbf{break}
        \EndIf
        \State $\boldsymbol{s} \gets $ \Call{$\operatorname{append}$}{$\boldsymbol{s}$, $s$}
      \EndFor
      \State \textbf{return} $\boldsymbol{s}$
    \EndFunction
  \end{algorithmic}
\end{algorithm}

The sampling procedure is described in generality by
\Cref{alg:llm-sequence-sampling}.  Often called multinomial sampling, the
procedure recursively generates new tokens by sampling from the categorical
distribution defined above until the \texttt{EOS} token is found.

\subsection{Guiding generation}
\label{sec:orge61fd8b}

We can derive other random variables from the next-token distribution by
manipulating the output logits \(\boldsymbol{\alpha}\).  Since we are dealing
with a finite, discrete distribution, we can compute an un-normalized
conditional distribution by applying a boolean mask \(m:
\mathcal{P}\left(\mathcal{V}\right) \to \left\{0, 1\right\}^N\) that restricts
the support of the original distribution:

\begin{align*}
  \boldsymbol{\alpha} &= \operatorname{LM}(\tilde{S}_t, \boldsymbol{\theta})\\
  \tilde{\boldsymbol{\alpha}} &= \operatorname{m}\left(\tilde{S}_t\right) \odot \boldsymbol{\alpha}\\
  \tilde{s}_{t+1} &\sim \operatorname{Categorical}({\tilde{\boldsymbol{\alpha}}})
\end{align*}

The resulting conditional distribution implied by \(\tilde{s}_{t+1}\) encodes
constraints on the support of \(s_{t+1}\). For instance, the masks \(m\) could
be designed so that the generated sequences, \(\tilde{S}_{t+1} =
\left(\tilde{s}_1, \dots, \tilde{s}_{t+1}\right)\), represent

\begin{itemize}
\item digit samples,
\item strings that match the regular expression \texttt{[a-zA-Z]},
\item and strings that parse according to a specified grammar (e.g. Python, SQL, etc.)
\end{itemize}

The sampling procedure with masking is a simple augmentation of
\Cref{alg:llm-sequence-sampling} and is provided in
\Cref{alg:llm-sequence-sampling-with-mask}.

\begin{algorithm}
  \caption{LLM token sampling with masking}
  \label{alg:llm-sequence-sampling-with-mask}
  \begin{algorithmic}[1]
    \Function{sample\_tokens}{$L$}
      \State $\boldsymbol{s} \gets () $
      \For{$i \gets 1, L$}
        \State $\boldsymbol{\alpha} \gets $
          \Call{$\operatorname{LLM}$}{$\boldsymbol{s}$, $\boldsymbol{\theta}$}
        \State Construct the mask $\operatorname{m}\left(\boldsymbol{s}\right)$ \label{construct-m}
        \State $\tilde{\boldsymbol{\alpha}} \gets \operatorname{m} \odot \boldsymbol{\alpha}$
        \State Sample $\tilde{s} \sim \operatorname{Categorical}({\tilde{\boldsymbol{\alpha}}})$
        \If{$\tilde{s} = \texttt{EOS}$}
          \State \textbf{break}
        \EndIf
        \State $\boldsymbol{s} \gets $ \Call{$\operatorname{append}$}{$\boldsymbol{s}$, $\tilde{s}$}
      \EndFor
      \State \textbf{return} $\boldsymbol{s}$
    \EndFunction
  \end{algorithmic}
\end{algorithm}

The computation of \(m\) on line
\(\algref{alg:llm-sequence-sampling-with-mask}{construct-m}\) is implicitly
performed over all the elements of \(\mathcal{V}\).  Aside from computing
\(\boldsymbol{\alpha}\), this step is easily the most expensive.  In the case of
regular expression-guided masking--and cases more sophisticated than that--the
support and, thus, \(m\) will necessarily depend on the previously sampled
tokens.  Guided generation of this kind is ultimately an iterative matching
or parsing problem and is not directly amenable to standard approaches
that require access to a complete string upfront.  In some cases, partial
matching or parsing can be performed from the start of the sampled sequence
on each iteration, but this has a cost that grows at least linearly
alongside the \(\mathcal{O}(N)\) cost of its application across the entire vocabulary.

This leads us to the main questions of this work: how can we efficiently match
or parse incomplete strings according to a regular expression or CFG and determine the masks
\(m\) at each iteration of \Cref{alg:llm-sequence-sampling-with-mask}?

\section{Iterative FSM Processing and Indexing}
\label{sec:iterative-fsm}
We frame the case of regular expression guided generation in terms of state machines.
This framing allows us to specify exactly how regular expression matching can be
arbitrarily started and stopped, so that it can be easily and efficiently
continued between samples of \(\tilde{s}_{i+1}\), as well as how the masks can
be computed without run-time evaluations over \(\mathcal{V}\).

To be precise, we consider regular expressions in 5-tuple finite automaton form
\citep[Definition 1.5]{SipserIntroductionTheoryComputation1996}:

\begin{definition}[Finite Automaton]
A finite automaton, or finite-state machine, is given by \((Q, \Sigma, \delta,
q_0, F)\), where \(Q\) is a finite set of states, \(\Sigma\) a finite alphabet,
\(\delta: Q \times \Sigma \to Q\) the transition function, \(q_0 \in Q\) the
start state, and \(F \subseteq Q\) the set of accept states.
\label{finite-automaton}
\end{definition}

The characters comprising the strings in \(\mathcal{V}\) are drawn from
\(\Sigma\): i.e. \(\mathcal{V} \subset \mathcal{P}(\Sigma)\).  Throughout, the
FSM states, \(Q\), will be represented by integer values for simplicity.

This formulation allows us to determine the exact states in \(Q\) in which the
guiding regular expression's FSM stops after sampling a single vocabulary token
\(\tilde{s}_{t+1}\).  These FSM states can then be tracked during the LLM
token sampling process in \Cref{alg:llm-sequence-sampling-with-mask} and used to
efficiently continue the state machine without reading from the beginning
of the growing sample sequence each time.

\begin{figure}[htbp]
\centering
\includegraphics[width=.9\linewidth]{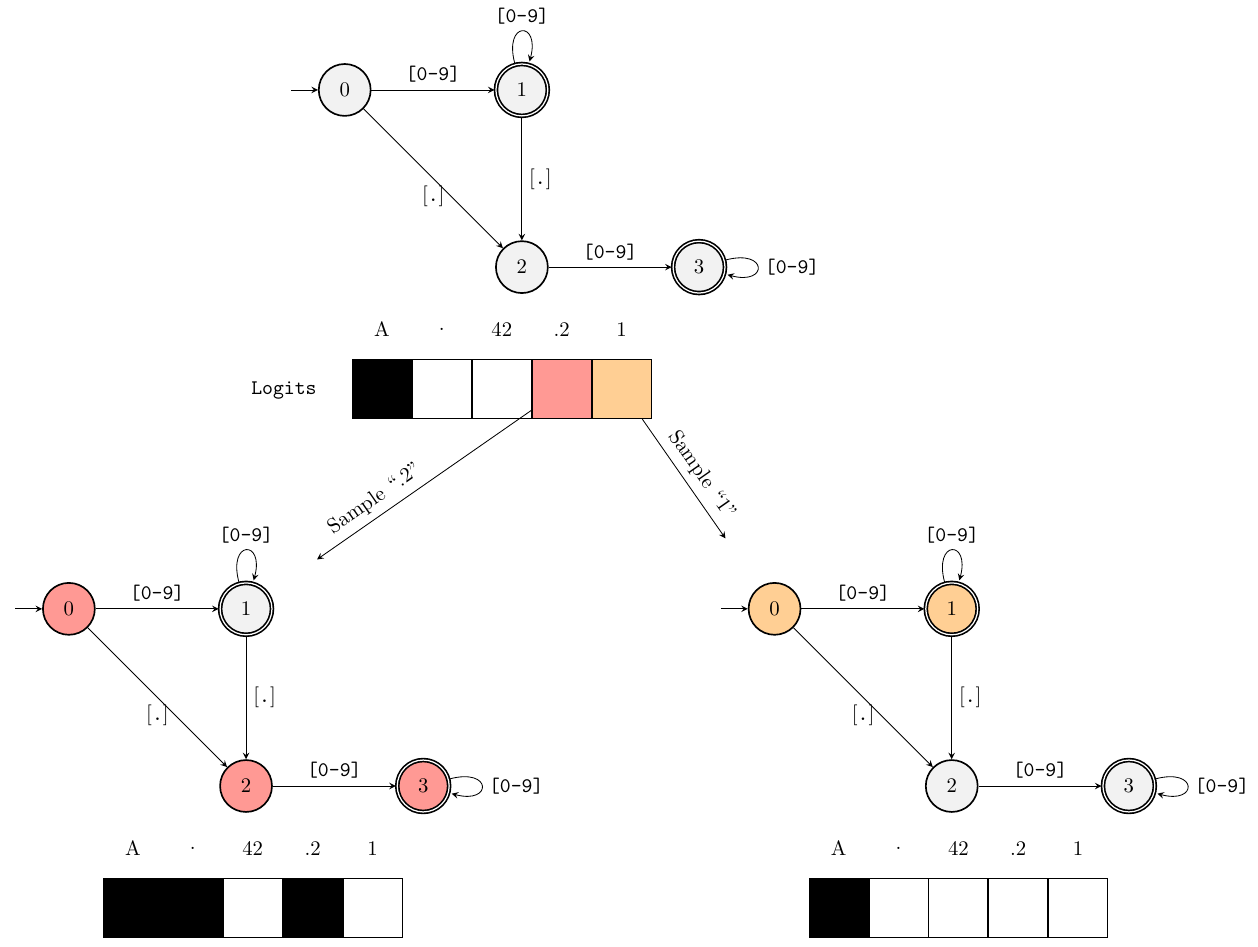}
\caption{\label{fig:fsm-logits-mask}FSM masking for the regular expression \texttt{([0-9]*)?\textbackslash{}.?[0-9]*}.}
\end{figure}

\begin{exa}
We illustrate the FSM sampling process in \Cref{fig:fsm-logits-mask} for the
regular expression \texttt{([0-9]*)?\textbackslash{}.?[0-9]*}, which can be used to generate
floating-point numbers. For simplicity, let the vocabulary, \(\mathcal{V}\),
consist of only the strings: \mintinline{python}{"A"}, \mintinline{python}{"."}, \mintinline{python}{"42"}, \mintinline{python}{".2"}, and \mintinline{python}{"1"}.

When the generation begins, the FSM is in state \texttt{0}, so our algorithm masks the
string \mintinline{python}{"A"}, since it would not be accepted by the FSM.  We can only
sample \mintinline{python}{"."}, \mintinline{python}{"42"}, \mintinline{python}{".2"}, and \mintinline{python}{"1"} in this case.

If we sample \mintinline{python}{".2"}, we advance the FSM to state \texttt{3}. In this case, only \mintinline{python}{"42"}
and \mintinline{python}{"1"} are valid completions, so we mask the other values before sampling. If
we sample \mintinline{python}{"1"} instead, we advance the FSM to state \texttt{1}, in which
case \mintinline{python}{"."}, \mintinline{python}{".42"}, \mintinline{python}{".2"}, and \mintinline{python}{"1"} are valid completions and the mask
remains unchanged.
\end{exa}

Looping through the vocabulary to determine the valid next tokens is still the
biggest issue.  For that, we pre-process the vocabulary using the regular
expression's FSM and build an index.  The important part is that we consider
starting in every viable FSM state, because the strings in the vocabulary could
match arbitrary parts of a regular expression, and those parts are implicitly
the FSM states.

A procedure for producing matches starting at any point in the FSM is given in
\Cref{alg:fsm-sub-sequences}.  The result is a list of sub-sequences detailing the
states through which the FSM would traverses when accepting the provided string.

\begin{algorithm}
  \caption{Find sub-sequences of the FSM $M$ that accept the string $\boldsymbol{v}$}
  \label{alg:fsm-sub-sequences}
  \begin{algorithmic}[1]
    \Function{find\_sub\_sequences}{$M$, $\boldsymbol{v}$}
      \State $M = (Q, \Sigma, \delta, q_0, F)$
      \State $res \gets ()$
      \For{$r \in \delta^{-1}(\cdot, v_0)$} \label{delta-inv}
        \Comment{Loop through states that read $v_0$}
        \State $p \gets (r)$
        \For{$i \gets 1, \abs{\boldsymbol{v}} - 1$}
          \Comment{Walk the FSM}
          \If{$\delta(r, v_i) = \emptyset$}
            \Comment{The FSM does not read $v_i$}
            \State $p \gets ()$
            \State \textbf{break}
            \Comment{Stop walking and try the next start state}
          \EndIf
          \State $r \gets \delta(r, v_i)$
          \State $p \gets$ \Call{$\operatorname{append}$}{$p$, $r$}
        \EndFor
        \State $res \gets$ \Call{$\operatorname{append}$}{$res$, $p$}
      \EndFor
      \State \textbf{return} $res$
    \EndFunction
  \end{algorithmic}
\end{algorithm}

By matching the starting states of these sub-sequences to the last FSM state
arrived at in a single step of the loop in
\Cref{alg:llm-sequence-sampling-with-mask}, we can efficiently index the
vocabulary with a map, \(\sigma: Q \to \mathcal{P}(\mathcal{V})\),
connecting FSM states and sets of elements of the vocabulary that will be
accepted by the FSM in those states.

\Cref{alg:fsm-index-map} describes the construction of \(\sigma\).
\begin{algorithm}
\caption{Construct a map from FSM states to subsets of \(\mathcal{V}\)}\label{alg:fsm-index-map}
  \begin{algorithmic}[1]
    \Function{map\_states\_to\_vocab}{$M$, $\mathcal{V}$}
      \State $M = (Q, \Sigma, \delta, q_0, F)$
      \State Initialize the map $\sigma$ with empty sets for each element in $Q$
      \For{$v \in \mathcal{V}$}
        \Comment{Loop through the vocabulary}
        \State $Z \gets$ \Call{$\operatorname{find\_sub\_sequences}$}{$M$, $v$}
        \For{$z \in Z$}
          \Comment{Loop through state sequences accepting $v$}
          \State $\sigma(z_0) \gets \sigma(z_0) \cup v $
        \EndFor
      \EndFor
      \State \textbf{return} $\sigma$
    \EndFunction
  \end{algorithmic}
\end{algorithm}

Using a hash-map for \(\sigma\) can make the \(m\) step in
\Cref{alg:llm-sequence-sampling-with-mask} cost only \(\mathcal{O}(1)\) on average.
Furthermore, since \(\sigma\) is constructed outside of the token sampling
procedure, its run-time cost is effectively irrelevant, although it
theoretically requires memory equal to the number of states in the FSM
(i.e. \(\abs{Q}\)).  Fortunately, for non-pathological combinations of regular
expressions and vocabularies, not every string in the vocabulary will be
accepted by the FSM, and not every FSM state will be represented by a string in
\(\mathcal{V}\).

\subsection{Examples}
\label{sec:org2f3c210}

In this section we use GPT2-medium (355M parameters) to illustrate how regular
expression guided generation works in practice. We use the library Outlines to
generate them:

\begin{oxtcblisting}{minted language=python, nofloat}
import outlines.models as models
import outlines.text.generate as generate

model = models.transformers("gpt2-medium")

prompt = "Is 1+1=2? "

unguided = generate.continuation(model, max_tokens=30)(prompt)
guided = generate.regex(model, r"\s*([Yy]es|[Nn]o|[Nn]ever|[Aa]lways)", max_tokens=30)(
    prompt
)

print(unguided)
# Is 1+1=2?
#
# This is probably the most perplexing question. As I said in one of my articles describing how I call 2 and 1, there isn't

print(guided)
# Is 1+1=2? Always

\end{oxtcblisting}

\begin{oxtcblisting}{minted language=python, nofloat}
prompt = "In what year was Noam Chomsky born?\n"
unguided = generate.continuation(model, max_tokens=30)(prompt)
guided = generate.regex(model, r"\s*19[0-9]{2}", max_tokens=30)(prompt)

print(unguided)
# In what year was Noam Chomsky born?
#
# Professor Chomsky was born in about 1895 in Mille Medad, near Paris. Like others Chomsky does not know the details of the birth weight of

print(guided)
# In what year was Noam Chomsky born?1952

\end{oxtcblisting}

\begin{oxtcblisting}{minted language=python, nofloat}
prompt = "What is the IP address of the Google DNS servers? "
unguided = generate.continuation(model, max_tokens=30)(prompt)
guided = generate.regex(
    model,
    r"((25[0-5]|2[0-4]\d|[01]?\d\d?)\.){3}(25[0-5]|2[0-4]\d|[01]?\d\d?)",
    max_tokens=30,
)(prompt)

print(unguided)
# What is the IP address of the Google DNS servers?
#
# Passive DNS servers are at DNS servers that are private. In other words, both IP servers are private. The database does not contain Chelsea Manning

print(guided)
# What is the IP address of the Google DNS servers?
# 2.2.6.1

\end{oxtcblisting}

\subsection{Comparison with current methods}
\label{sec:orgf11a9c9}
To illustrate the efficiency of the indexing approach described here, and implemented in
Outlines, we perform a simple comparison with the Guidance library.  As of this
writing, the Guidance library uses partial regular expression matching--applied
from the start of the sampled sequence each time--and must iterate over the LLM's
vocabulary (\(N = 50,257\)) on each step.

The Guidance code and prompt used for this comparison are as follows:
\begin{oxtcblisting}{minted language=python, nofloat}
import guidance

llm = guidance.llms.Transformers(
    "gpt2",
    token_healing=False,
    device="cuda",
    temperature=0.1,
)

program = guidance(
    f"""What is a good Python variable name?{{{{gen temperature=0.1 max_tokens={max_tokens} pattern="[^\W\d]\w*"}}}}""",
    llm=llm,
    caching=False,
    async_mode=False,
    stream=False,
    log=False,
    silent=True,
)

# Generate the token sequence.
# Only this call is timed.
program().text

\end{oxtcblisting}

The corresponding Outlines code is as follows:
\begin{oxtcblisting}{minted language=python, nofloat}
from outlines import disable_cache
import outlines.models as models
import outlines.text.generate as generate

disable_cache()

model = models.transformers("gpt2", device="cuda", temperature=0.1)

prompt = "What is a good Python variable name? "
guided_continuation = generate.regex(
    model,
    r"[^\W\d]\w*",
    max_tokens=max_tokens,
)


def reset_continuation():
    # This allows us to sample new sequences on each call
    guided_continuation.pstates = []
    return guided_continuation(prompt)


# Generate the token sequence.
# Only this call is timed.
reset_continuation()

\end{oxtcblisting}

The value of \mintinline{python}{max_tokens} is varied and the timings are recorded with \mintinline{python}{timeit}
for a single loop and single repeat value (i.e. only one sample is collected for each value
of \mintinline{python}{max_tokens}).  The results are plotted in \Cref{fig:plot-guidance-comparisons}.

Barring any configuration oversights that might be creating a large run-time
discrepancy, the observed scaling in the maximum number of sampled tokens is
striking and is indicative of the growing computational problem implied by the
approach.

\phantomsection
\label{fig:plot-guidance-comparisons}
\begin{center}
\includegraphics[width=.9\linewidth]{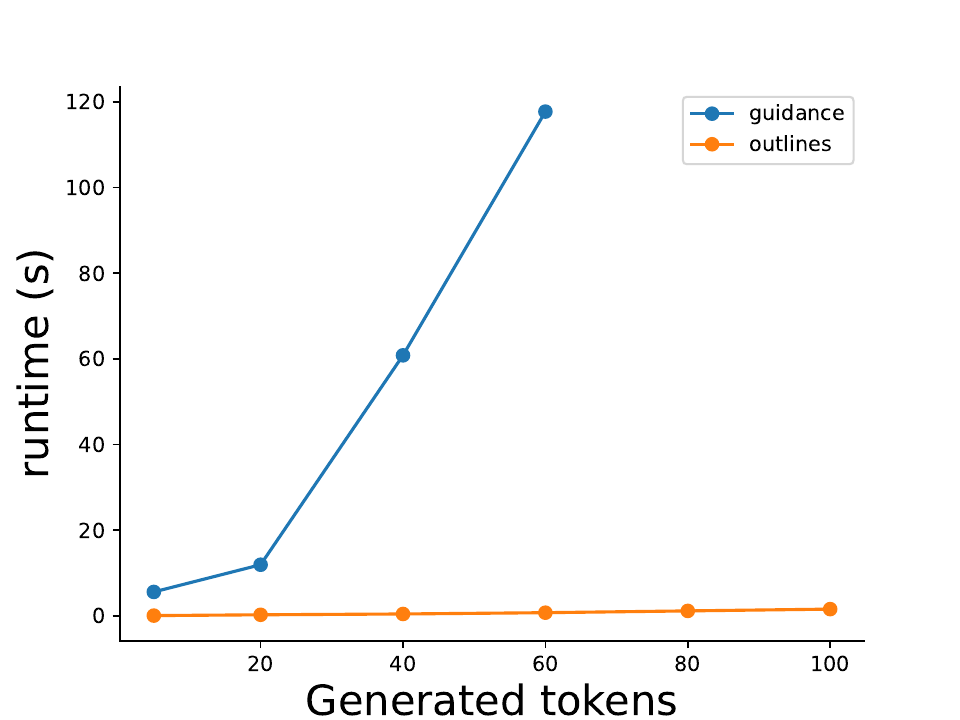}
\end{center}

\section{Extensions to Iterative Parsing}
\label{sec:parsing}
In this section, we move our focus to general parser-guided generation and
start with a simple walk-through for a Python-like grammar provided as a CFG.

Consider a vocabulary consisting of strings like \mintinline{python}{"d"} and \mintinline{python}{"ef"} that can be
combined to produce Python-like syntax according to an implicit CFG, and assume that
these strings are sequentially sampled and concatenated according to a process
like \Cref{alg:llm-sequence-sampling}.

Furthermore, consider a terminal symbol \texttt{DEF} in the CFG that corresponds to the
string \mintinline{python}{"def"} and is given by the trivial regular expression \texttt{def}.  Also,
consider a \texttt{NAME} symbol given by the regular expression \texttt{[\textasciicircum{}\textbackslash{}W\textbackslash{}d]\textbackslash{}w*}
(e.g. Python identifiers).  We want to sequentially parse strings sampled from
the aforementioned vocabulary in a way that adheres the Python syntax.

For example, the following could be one such
sequence: \mintinline{python}{["d", "ef", " f", "oo(", "):", " ", "pass"]}.  All the elements of the
sequence are by definition elements of the vocabulary.  Concatenating the sequence
produces \mintinline{python}{"def foo(): pass"}, which is a valid sequence of tokens defining a
function.  In the situation we're considering, we will have observed all the
tokens up to a certain point and know nothing about the ones after that point.

For instance, at the third observation in the example sequence, we have the
concatenated string \mintinline{python}{"def f"}.  If we were to lex/parse this string a
traditional approach would return the symbol sequence \texttt{DEF NAME}, which
misidentifies the \mintinline{python}{"f"} as a complete \texttt{NAME} token.  As we can see from the rest
of the sequence, the correct \texttt{NAME} token will be \mintinline{python}{"foo"}.

In general, the next valid strings that can be sampled from the vocabulary are
ones that either

\begin{enumerate}
\item continue expanding/advancing the \texttt{NAME} currently starting with \mintinline{python}{"f"} (as the
full sequence in our example does), and/or
\item anything that begins with \mintinline{python}{"("}--i.e. an \texttt{LPAR} symbol with regular expression
\texttt{(}--and proceeds to specify a valid argument signature.
\end{enumerate}

In the first case, the \mintinline{python}{"f"} can be seen as a partially matched \texttt{NAME} symbol in
Python, and--recalling that its regular expression is \texttt{[\textasciicircum{}\textbackslash{}W\textbackslash{}d]\textbackslash{}w*}--we can say
that it matches both sub-patterns (i.e. \texttt{[\textasciicircum{}\textbackslash{}W\textbackslash{}d]} and \texttt{\textbackslash{}w*}) in the regular
expression.  Our use of FSMs formalize the notion of sub-patterns by way of an
FSM's states.  In this case, the regex for \texttt{NAME} can be represented by an FSM,
\(M\), with three states: 0 (i.e. the initial state \(q_0\)), 1
(i.e. \texttt{[\textasciicircum{}\textbackslash{}W\textbackslash{}d]}), and 2 (i.e. \texttt{\textbackslash{}w*}), where \(1, 2 \in F\).

Using \Cref{alg:fsm-sub-sequences}, we would obtain the FSM state sequences \((0,
1)\), \((1, 2)\), \((2, 2)\) for \mintinline{python}{"f"} and the FSM, \(M\), corresponding to the \texttt{NAME}
symbol.  These FSM sequences for \mintinline{python}{"f"} tell us that matching can start for this
vocabulary string in the states 0, 1, or 2, and it can end in states 1 or 2.

According to case 1. above, parsing can be continued--for the \texttt{NAME}
symbol--after previously ending in states 1 or 2.  According to case 2., the
next string could also start with or contain an \texttt{LPAR}, implying that \(M\)
would have terminated, which it can given that 1 and 2 are final states in
\(M\) at which the parsing would have stopped after reading \mintinline{python}{"f"}.
\(M\) terminating also indicates that a \texttt{NAME} symbol was completed, and
that a transition to a state accepting \texttt{LPAR} was allowed by the grammar.

In this illustration, the next valid vocabulary strings are at
least \mintinline{python}{"d", "ef", "pass", " ", "oo("}, because all of those strings would expand the
partially matched \texttt{NAME}, and the last one would also progress the parse state
to one that reads an \texttt{LPAR}.  The remaining string, \mintinline{python}{"):"}, from the subset of
the vocabulary we've considered would result in a sequence with invalid syntax.

In relation to the FSM indexing approach, this means that \Cref{alg:fsm-index-map}
would map FSM states 0, 1, and 2 to the subset \mintinline{python}{"d", "ef", "pass", " ", "oo("}
for the symbol \texttt{NAME} and its FSM, \(M\).

This illustration omits the underlying parser states that determine which
grammar symbols and transitions are allowed.  We use pushdown automata (PDA) as
a means to extend the FSM approach and address the remaining details.

\subsection{Pushdown Automata Formulation}
\label{sec:org2e51496}
We define pushdown automata using the following 6-tuple
representation \citep[Definition 2.13]{SipserIntroductionTheoryComputation1996}:

\begin{definition}[Pushdown Automaton]
A pushdown automaton is given by \((Q, \Sigma, \Gamma, \delta, q_0, F)\), where
\(Q\), \(\Sigma\), \(\Gamma\), and \(F\) are all finite sets,
\(\Gamma\) is the stack alphabet,
\(\delta: Q \times \Sigma_\epsilon \times \Gamma_\epsilon \to \mathcal{P}\left(Q \times \Gamma_\epsilon\right)\),
\(\Gamma_\epsilon \equiv \Gamma \cup \epsilon\), \(\epsilon\) is the empty character,
and the remaining symbols retain their meanings from the finite automaton definition.
\label{pushdown-automaton}
\end{definition}

In order to construct an indexing approach for a PDA-driven parser, we need to
use the connection between a CFG's symbols--via a corresponding PDA's alphabet--and
the lexing and scanning steps that produce the symbols read by a PDA.

More specifically, parsers are supported by lexers and scanners that identify
symbols from a sequence of character inputs, as we implicitly illustrated in
\Cref{sec:parsing}.
Ordered lists of terminal symbols can be constructed for each parse/PDA state
based on the symbol and stack transitions allowed by the map \(\delta\) in each
state.  This means that we can construct an FSM for each parse state that is the
union of each FSM corresponding to a terminal symbols read by the state.

A scanning step will then identify a set of possible terminal symbols \(V \subset
\Sigma\) for the characters read since the last fully identified symbol in the
parsing process.  For example, in the initial state \(q_0\) of a PDA for the
Python-like CFG in \Cref{sec:parsing}, scanning and lexing the string \mintinline{python}{"de"} will
result in \(V = \left\{ \texttt{DEF}, \texttt{NAME} \right\}\): i.e. \texttt{DEF} for
any vocabulary string completing the string \mintinline{python}{"def"}--followed by a string not
also read by the \texttt{NAME} FSM (e.g. \mintinline{python}{"def "})--and \texttt{NAME} for any other strings
read by its FSM (e.g. \mintinline{python}{"default"}).
Note that steps of the scanner--and sampling steps of the LLM--will
eventually reduce the set \(V\) until a single terminal symbol \(v \in V\) is
determined.

By applying \Cref{alg:fsm-sub-sequences} to each string in \(\mathcal{V}\) using
the combined FSMs for each parse state, we can determine parser configurations that
consist of the PDA states, the corresponding FSM states, and the potential
terminal symbols.

By analogy with the steps in \Cref{alg:fsm-sub-sequences},
we can use the pre-image of the PDA's transition map to determine PDA stack
values that will read the PDA states \(q \in Q\) and terminal symbol sets \(V\)
of a parser configuration:
\begin{equation*}
  \delta^{-1}(q, V, \cdot) \equiv
    \left\{
      g : \delta(q, v, g) \in \mathcal{P}\left(Q \times \Gamma_\epsilon\right),
        g \in \Gamma_\epsilon, v \in V
    \right\}.
\end{equation*}

The stack values provided by this map are needed in order to find paths--if
any--through the PDA that allow successful, complete parses of
each string in \(\mathcal{V}\) starting from their possible parser
configurations.  For parser state and terminal combinations that correspond to
\(\texttt{REDUCE}\) operations of an \(\texttt{LALR(1)}\) parser, these parser
configurations will consist of more than just the top-of-stack values in
\(\Gamma\); they will consist of sub-stacks corresponding to all valid prefixes for the
\(\texttt{REDUCE}\) operations entailed by a vocabulary string.
Ultimately, each parser configuration that permits a complete parse of a
vocabulary string is added as an entry in the index for the PDA, and, in
this case, the index will need to be a trie data structure in order to allow
queries against the parser's stack values.

\section{Discussion}
\label{sec:orgb3a5bd1}

The vocabulary indexing introduced in this paper removes a prohibitive
run-time scaling barrier in guided generation.  Naturally, it makes a trade-off
between processing and memory, but we believe that the memory costs are relatively
low on average and--when not--can be reduced through conventional means.

In our tests using a slightly augmented version of the Python grammar, we find that
even naively constructed indices (i.e. ones containing unused and redundant
parser and FSM state configurations) are still only around 50 MB.  Furthermore,
these indices were constructed with un-reduced DFAs, implying that
there are numerous redundant states unnecessarily increasing the size of the
indices.  Likewise, if the exact representation of the state machines is ever an
issue, it's possible that other state machine formulations with lower memory
requirements could suffice (e.g. NFAs).

The implications of this work are not limited to neural text generation. For
instance, one could use the indexing approach described here to assist with the
\emph{training} or \emph{fine-tuning} of LLMs when structured outputs are
required.  We can also speculate that assisted generation during training may
reduce the need for a model to learn syntactic details.

In addition, this method provides an alternative way to evaluate current models.
One could, for instance, attempt to quantify the discrepancy between the masked
logits generated by our method and the raw logits generated by the model. Which
could in turn inform the training objective of a model.

It may also be possible to ``lift'' the masks computed by this approach into the
language models themselves.  Basically, the masks implicitly determine which
computations do \emph{not} need to be performed.  Our current formulation only
applies the masks at the lowest level, but, by lifting the masks
further up into the architecture of the model, we may be able to modulate
which slices of the model parameters are needed \emph{before} unnecessarily
performing operations on them.  This has the potential to further reduce
computational costs.

\bibliographystyle{plainnat}
\bibliography{efficient-guided-generation}

\begin{thebibliography}{19}
\providecommand{\natexlab}[1]{#1}
\providecommand{\url}[1]{\texttt{#1}}
\expandafter\ifx\csname urlstyle\endcsname\relax
  \providecommand{\doi}[1]{doi: #1}\else
  \providecommand{\doi}{doi: \begingroup \urlstyle{rm}\Url}\fi

\bibitem[{Beurer-Kellner} et~al.(2023){Beurer-Kellner}, Fischer, and
  Vechev]{Beurer-KellnerPromptingprogrammingquery2023}
Luca {Beurer-Kellner}, Marc Fischer, and Martin Vechev.
\newblock Prompting is programming: {{A}} query language for large language
  models.
\newblock \emph{Proceedings of the ACM on Programming Languages}, 7\penalty0
  (PLDI):\penalty0 1946--1969, 2023.

\bibitem[Dong et~al.(2023)Dong, Li, and Jin]{DongCODEPGrammaticalSeq2Seq2023}
Yihong Dong, Ge~Li, and Zhi Jin.
\newblock {{CODEP}}: {{Grammatical Seq2Seq Model}} for {{General-Purpose Code
  Generation}}.
\newblock In \emph{Proceedings of the 32nd {{ACM SIGSOFT International
  Symposium}} on {{Software Testing}} and {{Analysis}}}, {{ISSTA}} 2023, pages
  188--198, {New York, NY, USA}, July 2023. {Association for Computing
  Machinery}.
\newblock ISBN 9798400702211.
\newblock \doi{10.1145/3597926.3598048}.

\bibitem[Geng et~al.(2023)Geng, Josifosky, Peyrard, and
  West]{GengFlexibleGrammarBasedConstrained2023}
Saibo Geng, Martin Josifosky, Maxime Peyrard, and Robert West.
\newblock Flexible {{Grammar-Based Constrained Decoding}} for {{Language
  Models}}, May 2023.

\bibitem[Kuchnik et~al.(2023)Kuchnik, Smith, and
  Amvrosiadis]{KuchnikValidatinglargelanguage2023}
Michael Kuchnik, Virginia Smith, and George Amvrosiadis.
\newblock Validating large language models with relm.
\newblock \emph{Proceedings of Machine Learning and Systems}, 5, 2023.

\bibitem[Lew et~al.(2023)Lew, {Zhi-Xuan}, Grand, and
  Mansinghka]{LewSequentialMonteCarlo2023}
Alexander~K. Lew, Tan {Zhi-Xuan}, Gabriel Grand, and Vikash~K. Mansinghka.
\newblock Sequential {{Monte Carlo Steering}} of {{Large Language Models}}
  using {{Probabilistic Programs}}.
\newblock \emph{arXiv preprint arXiv:2306.03081}, 2023.

\bibitem[Louf and Willard()]{LoufoutlinesGenerativeModel}
R{\'e}mi Louf and Brandon~T. Willard.
\newblock Outlines: {{Generative Model Programming}}.
\newblock URL \url{https://github.com/normal-computing/outlines}.

\bibitem[{Microsoft}(2023)]{Microsoftguidance2023}
{Microsoft}.
\newblock Guidance.
\newblock Microsoft, July 2023.
\newblock URL \url{https://github.com/microsoft/guidance}.

\bibitem[Poesia et~al.(2022{\natexlab{a}})Poesia, Polozov, Le, Tiwari, Soares,
  Meek, and Gulwani]{PoesiaSynchromeshReliablecode2022}
Gabriel Poesia, Oleksandr Polozov, Vu~Le, Ashish Tiwari, Gustavo Soares,
  Christopher Meek, and Sumit Gulwani.
\newblock Synchromesh: {{Reliable}} code generation from pre-trained language
  models.
\newblock \emph{arXiv preprint arXiv:2201.11227}, 2022{\natexlab{a}}.

\bibitem[Poesia et~al.(2022{\natexlab{b}})Poesia, Polozov, Le, Tiwari, Soares,
  Meek, and Gulwani]{PoesiaSynchromeshReliablecode2022a}
Gabriel Poesia, Oleksandr Polozov, Vu~Le, Ashish Tiwari, Gustavo Soares,
  Christopher Meek, and Sumit Gulwani.
\newblock Synchromesh: {{Reliable}} code generation from pre-trained language
  models, January 2022{\natexlab{b}}.

\bibitem[Rabinovich et~al.(2017)Rabinovich, Stern, and
  Klein]{RabinovichAbstractsyntaxnetworks2017}
Maxim Rabinovich, Mitchell Stern, and Dan Klein.
\newblock Abstract syntax networks for code generation and semantic parsing.
\newblock \emph{arXiv preprint arXiv:1704.07535}, 2017.

\bibitem[Radford et~al.(2019)Radford, Wu, Child, Luan, Amodei, and
  Sutskever]{RadfordLanguagemodelsare2019}
Alec Radford, Jeffrey Wu, Rewon Child, David Luan, Dario Amodei, and Ilya
  Sutskever.
\newblock Language models are unsupervised multitask learners.
\newblock \emph{OpenAI blog}, 1\penalty0 (8):\penalty0 9, 2019.

\bibitem[Rickard(2023{\natexlab{a}})]{RickardparserLLM2023}
Matt Rickard.
\newblock {{parserLLM}}, July 2023{\natexlab{a}}.
\newblock URL \url{https://github.com/r2d4/parserllm}.

\bibitem[Rickard(2023{\natexlab{b}})]{Rickardr2d4rellmExact2023}
Matt Rickard.
\newblock R2d4/rellm: {{Exact}} structure out of any language model
  completion., 2023{\natexlab{b}}.
\newblock URL \url{https://github.com/r2d4/rellm}.

\bibitem[Scholak et~al.(2021)Scholak, Schucher, and
  Bahdanau]{ScholakPICARDParsingincrementally2021}
Torsten Scholak, Nathan Schucher, and Dzmitry Bahdanau.
\newblock {{PICARD}}: {{Parsing}} incrementally for constrained auto-regressive
  decoding from language models.
\newblock \emph{arXiv preprint arXiv:2109.05093}, 2021.

\bibitem[Sennrich et~al.(2015)Sennrich, Haddow, and
  Birch]{SennrichNeuralmachinetranslation2015}
Rico Sennrich, Barry Haddow, and Alexandra Birch.
\newblock Neural machine translation of rare words with subword units.
\newblock \emph{arXiv preprint arXiv:1508.07909}, 2015.

\bibitem[Sipser(1996)]{SipserIntroductionTheoryComputation1996}
Michael Sipser.
\newblock \emph{Introduction to the {{Theory}} of {{Computation}}}.
\newblock {International Thomson Publishing}, 1996.

\bibitem[Vaswani et~al.(2017)Vaswani, Shazeer, Parmar, Uszkoreit, Jones, Gomez,
  Kaiser, and Polosukhin]{VaswaniAttentionallyou2017}
Ashish Vaswani, Noam Shazeer, Niki Parmar, Jakob Uszkoreit, Llion Jones,
  Aidan~N. Gomez, {\textbackslash}Lukasz Kaiser, and Illia Polosukhin.
\newblock Attention is all you need.
\newblock \emph{Advances in neural information processing systems}, 30, 2017.

\bibitem[Wang et~al.(2023)Wang, Wang, Wang, Cao, Saurous, and
  Kim]{WangGrammarPromptingDomainSpecific2023}
Bailin Wang, Zi~Wang, Xuezhi Wang, Yuan Cao, Rif~A. Saurous, and Yoon Kim.
\newblock Grammar {{Prompting}} for {{Domain-Specific Language Generation}}
  with {{Large Language Models}}, May 2023.

\bibitem[Weng(2021)]{WengControllableNeuralText2021}
Lilian Weng.
\newblock Controllable {{Neural Text Generation}}, January 2021.
\newblock URL
  \url{https://lilianweng.github.io/posts/2021-01-02-controllable-text-generation/}.

\end{thebibliography}
\section*{Acknowledgments}
\label{sec:org89cb988}
We would like to thank Dan Gerlanc and Dan Simpson for their support and
constructive feedback.
\end{document}